\documentclass{article}
\usepackage[preprint]{nips_2018} 
\usepackage[utf8]{inputenc} 
\usepackage[T1]{fontenc}    
\usepackage{url}            
\usepackage{booktabs}       
\usepackage{nicefrac}       
\usepackage{microtype}      
\usepackage{xcolor}         
\usepackage{multirow}
\usepackage[pdftex]{graphicx}
\usepackage{amsmath}    
\usepackage{soul}
\usepackage{wrapfig}
\usepackage{subcaption}
\usepackage{array}
\usepackage{hyperref}
\usepackage{amsmath,amsfonts,amssymb,amsthm}
\usepackage{verbatim}
\definecolor{LinkColor}{rgb}{0.0, 0.18, 0.39}
\hypersetup{colorlinks=true,
    linkcolor=LinkColor,
    citecolor=LinkColor,
    filecolor=LinkColor,
    menucolor=LinkColor,
    urlcolor=LinkColor}

\author{
  Bryan McCann, Nitish Shirish Keskar, Caiming Xiong, Richard Socher\\
  Salesforce Research\\
  \texttt{\{bmccann,nkeskar,cxiong,rsocher\}@salesforce.com}
  }

\begin{document}
\newcommand{\challenge}{Natural Language Decathlon}
\newcommand{\chal}{decaNLP}
\newcommand{\Model}{Multitask Question Answering Network}
\newcommand{\model}{multitask question answering network}
\newcommand{\mdl}{MQAN}
\newcommand{\score}{decaScore}
\newcommand{\forward}{\rightarrow}
\newcommand{\backward}{\leftarrow}
\newcommand{\lstm}[2]{{\rm LSTM} \left( #1 , #2 \right)}
\newcommand{\bilstm}[1]{{\rm BiLSTM} \left( #1 \right)}
\newcommand{\sql}{^{\rm sql}}
\newcommand{\enc}{^{\rm enc}}
\newcommand{\dec}{^{\rm dec}}
\newcommand{\softmax}[1]{{\rm softmax} \left( #1 \right)}
\newcommand{\argmax}[1]{{\rm argmax} \left( #1 \right)}
\newcommand{\embin}[1]{L \left[ #1 \right]}
\newcommand{\embout}[1]{O \left[ #1 \right]}
\newcommand{\ftanh}[1]{{\rm tanh} \left( #1 \right)}
\newcommand{\nitish}[2]{{\color{red}\st{#1} #2}}
\newcommand{\nitishc}[2]{{\color{red}\textit{#1} \textbf{#2}}}
\newcommand{\gtxt}[1]{{\color{lightgray} {#1}}}

\title{The \challenge:\\ Multitask Learning as Question Answering}
\maketitle

\begin{abstract}
Deep learning has improved performance on many natural language processing (NLP) tasks individually.
However, general NLP models cannot emerge within a paradigm that focuses on the particularities of a single metric, dataset, and task.
We introduce the \challenge~(\chal), a challenge that spans ten tasks:
question answering, machine translation, summarization, natural language inference, sentiment analysis, semantic role labeling, relation extraction, goal-oriented dialogue, semantic parsing, and commonsense pronoun resolution.
We cast all tasks as question answering over a context.
Furthermore, we present a new \model~(\mdl) that jointly learns all tasks in \chal~without any task-specific modules or parameters.
\mdl~shows improvements in transfer learning for machine translation and named entity recognition, domain adaptation for sentiment analysis and natural language inference, and zero-shot capabilities for text classification.
We demonstrate that the {\mdl}'s multi-pointer-generator decoder is key to this success and that performance further improves with an anti-curriculum training strategy.
Though designed for decaNLP, MQAN also achieves state of the art results on the WikiSQL semantic parsing task in the single-task setting. 
We also release code for procuring and processing data, training and evaluating models, and reproducing all experiments for \chal.
\end{abstract}
\section{Introduction}
We introduce the \challenge~(\chal) in order to explore models that generalize to many different kinds of NLP tasks. \chal~encourages a single model to simultaneously optimize for ten tasks: question answering, machine translation, document summarization, semantic parsing, sentiment analysis, natural language inference, semantic role labeling, relation extraction, goal oriented dialogue, and pronoun resolution.
 
\begin{figure}[t!]
\centering
\includegraphics[width=\textwidth]{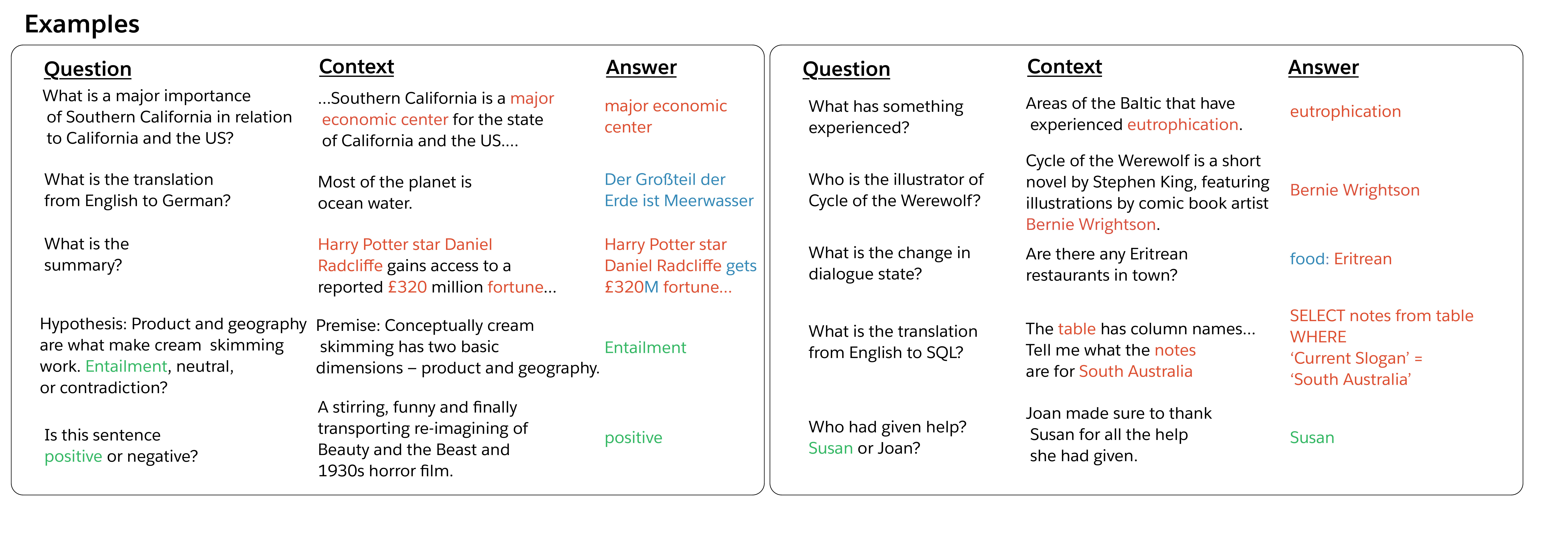}
\caption{\label{fig:one}
Overview of the decaNLP dataset with one example from each \chal~task in the order presented in Section~\ref{tasks}. They show how the datasets were pre-processed to become question answering problems. Answer words in red are generated by pointing to the context, in green from the question, and in blue if they are generated from a classifier over the output vocabulary. 
}
\vspace{-10pt}
\end{figure}

We frame all tasks as question answering~\citep{Kumar2016} by allowing task specification to take the form of a natural language question $q$: all inputs have a context, question, and answer (Fig.~\ref{fig:one}).
Traditionally, NLP examples have inputs $x$ and outputs $y$, and the underlying task $t$ is provided through explicit modeling constraints. 
Meta-learning approaches include $t$ as additional input~\citep{schmidhuber1987evolutionary,thrun1998learning,thrun1998lifelong,vilalta2002perspective}.
Our approach does not use a single representation for any $t$, but instead uses natural language questions that provide descriptions for underlying tasks. 
This allows single models to effectively multitask and makes them more suitable as pretrained models for transfer learning and meta-learning: natural language questions allow a model to generalize to completely new tasks through different but related task descriptions.

We provide a set of baselines for \chal~that combine the basics of sequence-to-sequence learning~\citep{Sutskever2014SequenceTS,Bahdanau2014NeuralMT,Luong2015EffectiveAT} with pointer networks~\citep{Vinyals2015,Merity2016PointerSM,Gulcehre2016,Gu2016,Nallapati2016AbstractiveTS}, advanced attention mechanisms~\citep{Xiong2016DynamicCN}, attention networks~\citep{Vaswani2017AttentionIA}, question answering ~\citep{Seo2017BidirectionalAF,Xiong2017DCNMO,Yu2016EndtoEndRC,Weissenborn2017MakingNQ}, and curriculum learning~\citep{Bengio2009CurriculumL}.

The \model~(\mdl) is designed for \chal~and makes use of a novel dual coattention and multi-pointer-generator decoder to multitask across all tasks in \chal.
Our results demonstrate that training the \mdl~jointly on all tasks with the right anti-curriculum strategy can achieve performance comparable to that of ten separate {\mdl}s, each trained separately. A \mdl~pretrained on \chal~shows improvements in transfer learning for machine translation and named entity recognition, domain adaptation for sentiment analysis and natural language inference, and zero-shot capabilities for text classification. Though not explicitly designed for any one task, \mdl~proves to be a strong model in the single-task setting as well, achieving state-of-the-art results on the semantic parsing component of \chal.

We have released code\footnote{\url{https://github.com/salesforce/decaNLP}} for obtaining and preprocessing datasets, training and evaluating models, and tracking progress through a leaderboard based on decathlon scores (\score). We hope that the combination of these resources will facilitate research in multitask learning, transfer learning, general embeddings and encoders, architecture search, zero-shot learning, general purpose question answering, meta-learning, and other related areas of NLP.

\section{Tasks and Metrics}\label{tasks}
\chal~consists of 10 publicly available datasets with examples cast as (question, context, answer) triplets as shown in Fig.~\ref{fig:one}.

\textbf{Question Answering.}
Question answering (QA) models receive a question and a context that contains information necessary to output the desired answer.
We use the Stanford Question Answering Dataset (SQuAD)~\citep{Rajpurkar2016SQuAD10} for this task. 
Contexts are paragraphs taken from the English Wikipedia, and answers are sequences of words copied from the context.
SQuAD uses a normalized F1 (nF1) metric that strip out articles and punctuation.

\textbf{Machine Translation.}
Machine translation models receive an input document in a source language that must be translated into a target language. We use the 2016 English to German training data prepared for the International Workshop on Spoken Language Translation (IWSLT)~\citep{Cettolo2016TheI2}. Examples are from transcribed TED presentations that cover a wide variety of topics with conversational language. We evaluate with a corpus-level BLEU score~\citep{Papineni2002BleuAM} on the 2013 and 2014 test sets as validation and test sets, respectively.

\textbf{Summarization.}
Summarization models take in a document and output a summary of that document.
Most important to recent progress in summarization was the transformation of the CNN/DailyMail (CNN/DM) corpus~\citep{Hermann2015TeachingMT} into a summarization dataset~\citep{Nallapati2016AbstractiveTS}.
We include the non-anonymized version of this dataset in \chal. On average, these examples contain the longest documents in \chal~and force models to balance extracting from the context with generation of novel, abstractive sequences of words.
CNN/DM uses ROUGE-1, ROUGE-2, and ROUGE-L scores~\citep{Lin2004ROUGEAP}.
We average these three measures to compute an overall ROUGE score.

\textbf{Natural Language Inference.}
Natural Language Inference (NLI) models receive two input sentences: a premise and a hypothesis. Models must then classify the inference relationship between the two as one of entailment, neutrality, or contradiction. 
We use the Multi-Genre Natural Language Inference Corpus (MNLI)~\citep{Williams2017ABC} which provides training examples from multiple domains (transcribed speech, popular fiction, government reports) and test pairs from seen and unseen domains. MNLI uses an exact match (EM) score.

\begin{table}[t]
\caption{\label{datasetInformation}
Summary of openly available benchmark datasets in \chal~and evaluation metrics that contribute to the \score. All metrics are case insensitive.
nF1 is the normalized F1 metric used by SQuAD that strips out articles and punctuation. EM is an exact match comparison: for text classification, this amounts to accuracy; for WOZ it is equivalent to turn-based dialogue state exact match (dsEM) and for WikiSQL it is equivalent to exact match of logical forms (lfEM). F1 for QA-ZRE is a corpus level metric (cF1) that takes into account that some question are unanswerable. Precision is the true positive count divided by the number of times the system returned a non-null answer. Recall is the true positive count divided by the number of instances that have an answer.}
\centering
\vspace{2pt}
\begin{tabular}{llrrrr}
\toprule
Task & Dataset  & \# Train & \# Dev & \# Test & Metric\\
         \midrule
Question Answering            & SQuAD    &$87599$ &$10570$& $9616$ & nF1     \\ 
Machine Translation           & IWSLT     &$196884$&$993$& $1305$ & BLEU    \\ 
Summarization                 & CNN/DM &$287227$ & $13368$ & $11490$ & ROUGE \\ 
Natural Language Inference    & MNLI      &$392702$& $20000$& $20000$ & EM     \\ 
Sentiment Analysis            & SST      &$6920$ &$872$&$1821$ & EM  \\ 
Semantic Role Labeling        & QA-SRL    &$6414$&$2183$& $2201$ & nF1     \\ 
Zero-Shot Relation Extraction & QA-ZRE   &$840000$ &$600$& $12000$ & cF1      \\ 
Goal-Oriented Dialogue        & WOZ      &$2536$ & $830$ & $1646$ & dsEM      \\ 
Semantic Parsing     & WikiSQL   &$56355$& $8421$& $15878$ & lfEM      \\ 
Pronoun Resolution            & MWSC      & $80$ & $82$ & $100$ & EM     \\ 
\bottomrule
\end{tabular}
\vspace{-15pt}
\end{table}

\textbf{Sentiment Analysis.}
Sentiment analysis models are trained to classify the sentiment expressed by input text.
The Stanford Sentiment Treebank (SST)~\citep{Socher2013EMNLP} consists of movie reviews with the corresponding sentiment (positive, neutral, negative).
We use the unparsed, binary version~\citep{Radford2017LearningTG}. SST also uses an EM score.

\textbf{Semantic Role Labeling.}
Semantic role labeling (SRL) models are given a sentence and predicate (typically a verb) and must determine `who did what to whom,' `when,' and `where'~\citep{Johansson2008DependencybasedSR}.
We use an SRL dataset that treats the task as question answering, QA-SRL~\citep{He2015QuestionAnswerDS}. This dataset covers both news and Wikipedia domains, but we only use the latter in order to ensure that all data for decaNLP can be freely downloaded. We evaluate QA-SRL with the nF1 metric used for SQuAD.

\textbf{Relation Extraction.} Relation extraction systems take in a piece of unstructured text and the kind of relation that is to be extracted from that text. In this setting, it is important that models can report that the relation is not present and cannot be extracted. As with SRL, we use a dataset that maps relations to a set of questions so that relation extraction can be treated as question answering: QA-ZRE~\citep{Levy2017ZeroShotRE}. Evaluation of the dataset is designed to measure zero shot performance on new kinds of relations -- the dataset is split so that relations seen at test time are unseen at train time. This kind of zero-shot relation extraction, framed as question answering, makes it possible to generalize to new relations. QA-ZRE uses a corpus-level F1 metric (cF1) in order to accurately account for unanswerable questions. This F1 metric defines precision as the true positive count divided by the number of times the system returned a non-null answer and recall as the true positive count divided by the number of instances that have an answer.

\textbf{Goal-Oriented Dialogue.} Dialogue state tracking is a key component of goal-oriented dialogue systems. Based on user utterances, actions taken already, and conversation history, dialogue state trackers keep track of which predefined goals the user has for the dialogue system and which kinds of requests the user makes as the system and user interact turn-by-turn. We use the English Wizard of Oz (WOZ) restaurant reservation task~\citep{wen2016network}, which comes with a predefined ontology of foods, dates, times, addresses, and other information that would help an agent make a reservation for a customer. WOZ is evaluated by turn-based dialogue state EM (dsEM) over the goals of the customers.

\textbf{Semantic Parsing.}
SQL query generation is related to semantic parsing. Models based on the WikiSQL dataset~\citep{zhongSeq2SQL2017} translate natural language questions into structured SQL queries so that users can interact with a database in natural language. WikiSQL is evaluated by a logical form exact match (lfEM) to ensure that models do not obtain correct answers from incorrectly generated queries.

\textbf{Pronoun Resolution.}
Our final task is based on Winograd schemas~\citep{winograd1972understanding}, which require pronoun resolution: "Joan made sure to thank Susan for the help she had [given/received]. Who had [given/received] help? Susan or Joan?".
We started with examples taken from the Winograd Schema Challenge~\citep{levesque2011winograd} and modified them to ensure that answers were a single word from the context. This modified Winograd Schema Challenge (MWSC) ensures that scores are neither inflated nor deflated by oddities in phrasing or inconsistencies between context, question, and answer. We evaluate with an EM score. 

\textbf{The Decathlon Score (\score).}
Models competing on \chal~are evaluated using an additive combination of each task-specific metric.  All metrics fall between $0$ and $100$, so that the \score~naturally falls between $0$ and $1000$ for ten tasks. Using an additive combination avoids issues that arise from weighing different metrics. All metrics are case insensitive.

\begin{figure}[t!]
\centering
\includegraphics[width=\textwidth]{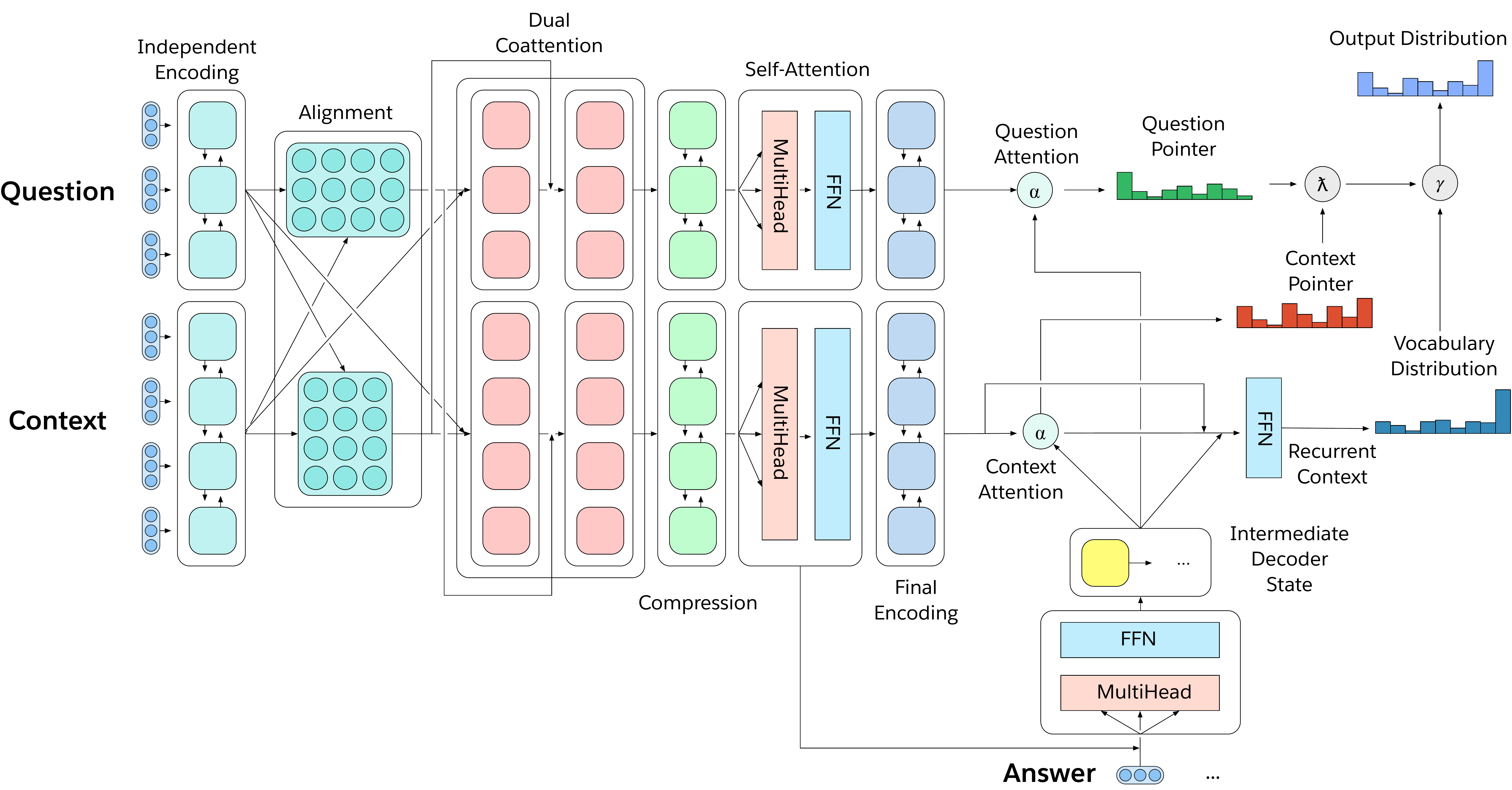}
\label{fig:model}
\caption{
Overview of the \mdl~model. It takes in a question and context document, encodes both with a BiLSTM, uses dual coattention to condition representations for both sequences on the other, compresses all of this information with another two BiLSTMs, applies self-attention to collect long-distance dependency, and then uses a final two BiLSTMs to get representations of the question and context. The multi-pointer-generator decoder uses attention over the question, context, and previously output tokens to decide whether to copy from the question, copy from the context, or generate from a limited vocabulary.}
\vspace{-15pt}
\end{figure}

\section{\Model~(\mdl)} 
\label{model}
Because every task is framed as question answering and trained jointly, we call our model a \model~(\mdl). 
Each example consists of a context, question, and answer as shown in Fig.~\ref{fig:one}.
Many recent QA models for question answering typically assume the answer can be copied from the context~\citep{wang2017machine,Seo2017BidirectionalAF,Xiong2017DCNMO}, but this assumption does not hold for general question answering. 
The question often contains key information that constrains the answer space. 
Noting this, we extend the coattention of \citep{Xiong2016DynamicCN} to enrich the representation of not only the input but also the question. Also, the pointer-mechanism of \citep{See2017GetTT} is generalized into a hierarchical, multi-pointer-generator that enables the capacity to copy directly from the question and the context.

During training, the \mdl~takes as input three sequences: a context $c$ with $l$ tokens, a question $q$ with $m$ tokens, and an answer $a$ with $n$ tokens. Each of these is represented by a matrix where the $i$th row of the matrix corresponds to a $d_{emb}$-dimensional embedding (such as word or character vectors) for the $i$th token in the sequence: \begin{equation}
C \in \mathbb{R}^{l \times d_{emb}}
\qquad 
Q \in \mathbb{R}^{m \times d_{emb}}
\qquad A \in \mathbb{R}^{n \times d_{emb}}
\end{equation}

An encoder takes these matrices as input and uses a deep stack of recurrent, coattentive, and self-attentive layers to produce final representations, $C_{fin} \in \mathbb{R}^{l \times d}$ and $Q_{fin}  \in \mathbb{R}^{m \times d}$, of both context and question sequences designed to capture local and global interdependencies. Appendix~\ref{sec:encoderDetails} describes the full details of the encoder.

\textbf{Answer Representations.} During training, the decoder begins by projecting the answer embeddings onto a $d$-dimensional space:
\begin{equation}
AW_2 = A_{proj} \in \mathbb{R}^{n \times d}
\end{equation}

This is followed by a self-attentive layers, which has a corresponding self-attentive layer in the encoder. Because it lacks both recurrence and convolution, we add to $A_{proj}$ positional encodings~\citep{Vaswani2017AttentionIA} $PE \in \mathbb{R}^{n \times d}$ with entries
\begin{equation}
PE[t,k] =
\begin{cases} 
\text{sin}(t/10000^{k/2d})  & \text{$k$ is even} \\
\text{cos}(t/10000^{(k-1)/2d}) & \text{$k$ is odd}   
\end{cases}
\qquad 
A_{proj} + PE = A_{ppr} \in \mathbb{R}^{n \times d}
\end{equation}

\textbf{Multi-head Decoder Attention.} We use self-attention\footnote{The decoder operates step by step. To prevent the decoder from seeing future time-steps during training, appropriate entries of $XY^\top$ are set to a large negative number prior to the softmax in~\eqref{attention}.}~\citep{Vaswani2017AttentionIA} so that the decoder is aware of previous outputs (or a special intialization token in the case of no previous outputs) and attention over the context to prepare for the next output. Refer to Appendix~\ref{sec:encoderDetails} for definitions of MultiHead attention and FFN, the residual feedforward network applied after MultiHead attention over the context.
\begin{equation}
\text{MultiHead}_A(A_{ppr}, A_{ppr}, A_{ppr}) = A_{mha} \in \mathbb{R}^{n \times d}
\end{equation}
\begin{equation}
\text{MultiHead}_AC((A_{mha} + A_{ppr}), C_{fin}, C_{fin}) = A_{ac} \in \mathbb{R}^{n \times d}
\end{equation}
\begin{equation}
FFN_A(A_{ac} + A_{mha} + A_{ppr}) = A_{self}  \in \mathbb{R}^{n \times d}
\end{equation}

\textbf{Intermediate Decoder State.} We next use a standard LSTM with attention to get a recurrent context state $\tilde c_{t}$ for time-step $t$. First, the LSTM produces an intermediate state $h_t$ using the previous answer word $ A_{self}^{t-1}$ and recurrent context state~\citep{Luong2015EffectiveAT}:
\begin{equation}
\text{LSTM}([ \left(A_{self}\right)_{t-1}; \tilde c_{t-1} ], {h_{t-1}}) = h_{t} \in \mathbb{R}^{d}
\end{equation}

\textbf{Context and Question Attention.} This intermediate state is used to get attention weights $\alpha_{t}^C$ and $\alpha_{t}^Q$ to allow the decoder to focus on encoded information relevant to time step $t$.
\begin{equation}
\softmax{C_{fin} (W_2 h_{t})} = \alpha_{t}^C \in \mathbb{R}^{l}
\qquad 
\softmax{Q_{fin} (W_3 h_{t})} = \alpha_{t}^Q \in \mathbb{R}^{m}
\end{equation}

\textbf{Recurrent Context State.} Context representations are combined with these weights and fed through a feedforward network with tanh activation to form the recurrent context state and question state:
\begin{equation}
 \ftanh{W_4 \left[ C_{fin}^\top \alpha_{t}^C; h_{t} \right]}  = \tilde c_{t}  \in \mathbb{R}^{d}
 \qquad
\ftanh{W_5 \left[ Q_{fin}^\top \alpha_{t}^Q; h_{t} \right]} = \tilde q_{t}  \in \mathbb{R}^{d}
\end{equation}

\textbf{Multi-Pointer-Generator.}
Our model must be able to generate tokens that are not in the context or the question. We give it access to $v$ additional vocabulary tokens. We obtain distributions over tokens in the context, question, and this external vocabulary, respectively, as 
\begin{equation}
\sum_{i:c_i=w_t} \left(\alpha_{t}^C \right)_i = p_c(w_t) \in \mathbb{R}^{n}
\qquad
\sum_{i:q_i=w_t} \left(\alpha_{t}^Q \right)_i  = p_q(w_t)\in \mathbb{R}^{m}
\end{equation}
\begin{equation}
\label{generativeVocabulary}
\softmax{W_v \tilde c_t} = p_v(w_t) \in \mathbb{R}^{v}
\end{equation}
These distributions are extended to cover the union of the tokens in the context, question, and external vocabulary by setting missing entries in each to $0$ so that each distribution is in $\mathbb{R}^{l+m+v}$. Two scalar switches regulate the importance of each distribution in determining the final output distribution.
\begin{equation}
\sigma \left ( W_{pv} \left [ \tilde c_t; h_{t};  \left(A_{self}\right)_{t-1} \right ] \right ) = \gamma \in \left [0,1 \right]
\qquad
\sigma \left ( W_{cq} \left [ \tilde q_t; h_{t}; \left(A_{self}\right)_{t-1} \right ] \right ) = \lambda \in \left [0,1 \right]
\end{equation}
\begin{equation}
\gamma p_v(w_t) + \left ( 1 - \gamma \right ) [\lambda p_c(w_t) + (1 - \lambda) p_q(w_t) ] = p(w_t) \in \mathbb{R}^{l+m+v}
\end{equation}

We train using a token-level negative log-likelihood loss over all time-steps: $\mathcal{L} =  -\sum_t^T \log p(a_t)$.
\section{Experiments and Analysis} 
\begin{table}[t]
\centering
\caption{Validation metrics for \chal~baselines: sequence-to-sequence (S2S) with self-attentive transformer layers (w/SAtt), the addition of coattention (+CAtt) over a split context and question, and a question pointer (+QPtr). The last model is equivalent to MQAN. Multitask models use a round-robin batch-level sampling strategy to jointly train on the full \chal. The last column includes an additional anti-curriculum (+ACurr) phase that trains on SQuAD alone before switching to the fully joint strategy. Entries marked with '-' would correspond to {\score}s for aggregates of separately trained models; this is not well-defined without a mechanism for choosing between models.
\label{modelAblations}
}
\begin{tabular}{lrrrrrrrrr}
\toprule
& \multicolumn{4}{c}{Single-task Training} & \multicolumn{4}{c}{Multitask Training} \\
\cmidrule[.5pt](lr){2-5}\cmidrule[.5pt](lr){6-10}
Dataset  & S2S & w/SAtt & +CAtt & +QPtr
         & S2S & w/SAtt & +CAtt & +QPtr & +ACurr
\\\midrule
SQuAD  & 48.2 & 68.2 & 74.6 & 75.5 & 47.5 & 66.8 & 71.8 & 70.8 & 74.3 \\  
IWSLT  & 25.0 & 23.3 & 26.0 & 25.5 & 14.2 & 13.6 & 9.0  & 16.1 & 13.7 \\  
CNN/DM & 19.0 & 20.0 & 25.1 & 24.0 & 25.7 & 14.0 & 15.7 & 23.9 & 24.6 \\ 
MNLI   & 67.5 & 68.5 & 34.7 & 72.8 & 60.9 & 69.0 & 70.4 & 70.5 & 69.2 \\  
SST    & 86.4 & 86.8 & 86.2 & 88.1 & 85.9 & 84.7 & 86.5 & 86.2 & 86.4 \\  
QA-SRL & 63.5 & 67.8 & 74.8 & 75.2 & 68.7 & 75.1 & 76.1 & 75.8 & 77.6 \\  
QA-ZRE & 20.0 & 19.9 & 16.6 & 15.6 & 28.5 & 31.7 & 28.5 & 28.0 & 34.7 \\ 
WOZ    & 85.3 & 86.0 & 86.5 & 84.4 & 84.0 & 82.8 & 75.1 & 80.6 & 84.1 \\  
WikiSQL& 60.0 & 72.4 & 72.3 & 72.6 & 45.8 & 64.8 & 62.9 & 62.0 & 58.7 \\ 
MWSC   & 43.9 & 46.3 & 40.4 & 52.4 & 52.4 & 43.9 & 37.8 & 48.8 & 48.4 \\ 
\midrule
\score &-&-&-&-&473.6&546.4&533.8&562.7&\textbf{571.7}\\
\bottomrule
\end{tabular}
\end{table}
\subsection{Baselines and MQAN}
In our framework, training examples are (question, context, answer) triplets.
Our first baseline is a pointer-generator sequence-to-sequence (S2S) model~\citep{See2017GetTT}. S2S models take in only a single input sequence, so we concatenate the context and question for this model. In Table~\ref{modelAblations}, validation metrics reveal that the S2S model does not perform well on SQuAD. On WikiSQL, it obtains a much higher score than prior sequence-to-sequence baselines~\citep{zhongSeq2SQL2017}, but it is low compared to MQAN (+QPtr) and the other baselines.

Augmenting the S2S model with self-attentive (w/ SAtt) encoder and decoder layers~\cite{Vaswani2017AttentionIA}, as detailed in~\ref{sec:encoderDetails}, increases the model's capacity to integrate information from both context and question. This improves performance on SQuAD by $20$ nF1, QA-SRL by $4$ nF1, and WikiSQL by $12$ LFEM. For WikiSQL, this model nearly matches the prior state-of-the-art validation results of $72.4\%$ without using a structured approach~\citep{dong2018coarse,huang2018natural,yu2018typesql}.

We next explore splitting the context and question into two input sequences and augmenting the S2S model with a coattention mechanism (+CAtt). Performance on SQuAD and QA-SRL increases by more than $5$ nF1 each. Unfortunately, this fails to improve other tasks, and it significantly hurts performance on MNLI and MWSC. For these two tasks, answers can be copied directly from the question. Because both S2S baselines had the question concatenated to the context, the pointer-generator mechanism was able to copy directly from the question. When the context and question were separated into two different inputs, the model lost this ability. 

To remedy this, we add a question pointer (+QPtr) to the previous baseline, which gives the \mdl~described in Section~\ref{model} and Appendix~\ref{sec:encoderDetails}. This boosts performance on both MNLI and MWSC above prior baselines. It also improved performance on SQuAD to $75.5$ nF1, which matches performance of the first wave of SQuAD models to make use of direct span supervision~\citep{Xiong2016DynamicCN}. This makes it the highest performing question answering model trained on SQuAD dataset that does not explicitly model the problem as span extraction. 

This last model achieved a new state-of-the-art test result on WikiSQL by reaching $72.4\%$ lfEM and $80.4\%$ database execution accuracy, surpassing the previous state of the art set by~\citep{dong2018coarse} at $71.7\%$ and $78.5\%$.

In the multitask setting, we see similar results, but we also notice several additional striking features. QA-ZRE performance increases $11$ F1 points over the highest single-task models, which supports the hypothesis that multitask learning can lead to better generalization for zero-shot learning.

Performance on tasks that require heavy use of the external vocabulary drops more than $50$ percent from the S2S baselines until the question pointer is added to the model. In addition to a coattended context, this question pointer makes use of a coattended question, which allows information from the question to flow directly into the decoder. We hypothesize that more direct access to the question makes it easier for the model decide when generating output tokens is more appropriate than copying. 

See Appendix~\ref{preprocessingTrainingDetails} for details regarding pre-processing and hyperparameters.

\subsection{Optimization Strategies and Curriculum Learning}   
For multitask training, we experiment with various round-robin batch-level sampling strategies. Fully joint training cycles through all tasks from the beginning of training. However, some tasks require more iterations to converge in the single-task setting, which suggests that these are more difficult for the model to learn. We experiment with both curriculum and anti-curriculum strategies~\cite{Bengio2009CurriculumL} based on this notion of difficulty.

We divide tasks into two groups: the easiest difficult task requires more than twice the iterations the most difficult easy task requires. Compared to the fully joint strategy, curriculum learning jointly trains the easier tasks (SST, QA-SRL, QA-ZRE, WOZ, WikiSQL, and MWSC) first. This leads to a dramatically reduced \score~(Appendix~\ref{appendix:curriculum}). Anti-curriculum strategies boost performance on tasks trained early, but can also hurt performance on tasks held out until later training. Of the various anti-curriculum strategies we experimented with, only the one which trains on SQuAD alone before transitioning to a fully joint strategy yielded a \score~higher than using the fully joint strategy without modification. For a full comparison, see Appendix~\ref{appendix:curriculum}.
\subsection{Analysis}
\begin{figure}
  \centering
 \includegraphics[width=\textwidth]{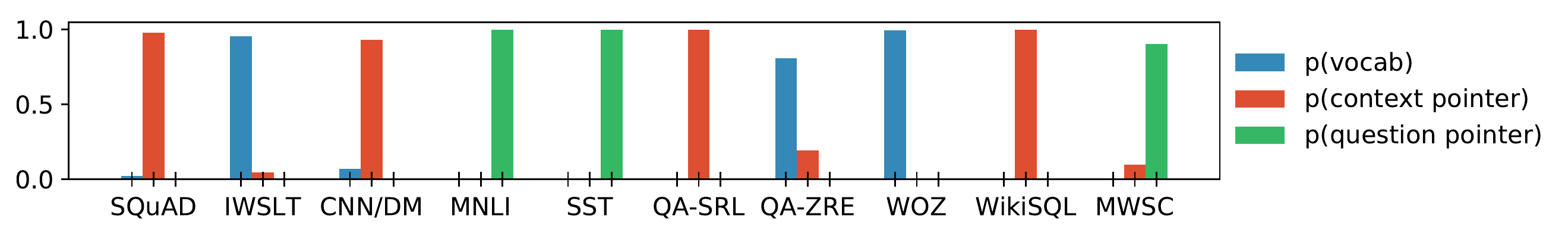}
  \caption{An analysis of how the \mdl~chooses to output answer words. When p(generation) is highest, the \mdl~places the most weight on the external vocab. When p(context) is highest, the \mdl~places the most weight on the pointer distribution over the context. When p(question) is highest, the \mdl~places the most weight on the pointer distribution over the question.
  }\label{pointerUsage}
\end{figure}
\textbf{Multi-Pointer-Generator and task identification.}
At each step, the \mdl~decides between three choices: generating from the vocabulary, pointing to the question, and pointing to the context. While the model is not trained with \textit{explicit} supervision for these decisions, it learns to switch between the three options. Fig.~\ref{pointerUsage} presents statistics of how often the final model chooses each option. For SQuAD, QA-SRL, and WikiSQL, the model mostly copies from the context. This is intuitive because all tokens necessary to correctly answer questions from these datasets are contained in the context. The model also usually copies from the context for CNN/DM because answer summaries consist mostly of words from the context with few words generated from outside the context in between. 

\begin{figure}
\centering
\begin{subfigure}[t]{0.45\textwidth}\label{fig:betterInitializationMT}
\includegraphics[width=\linewidth]{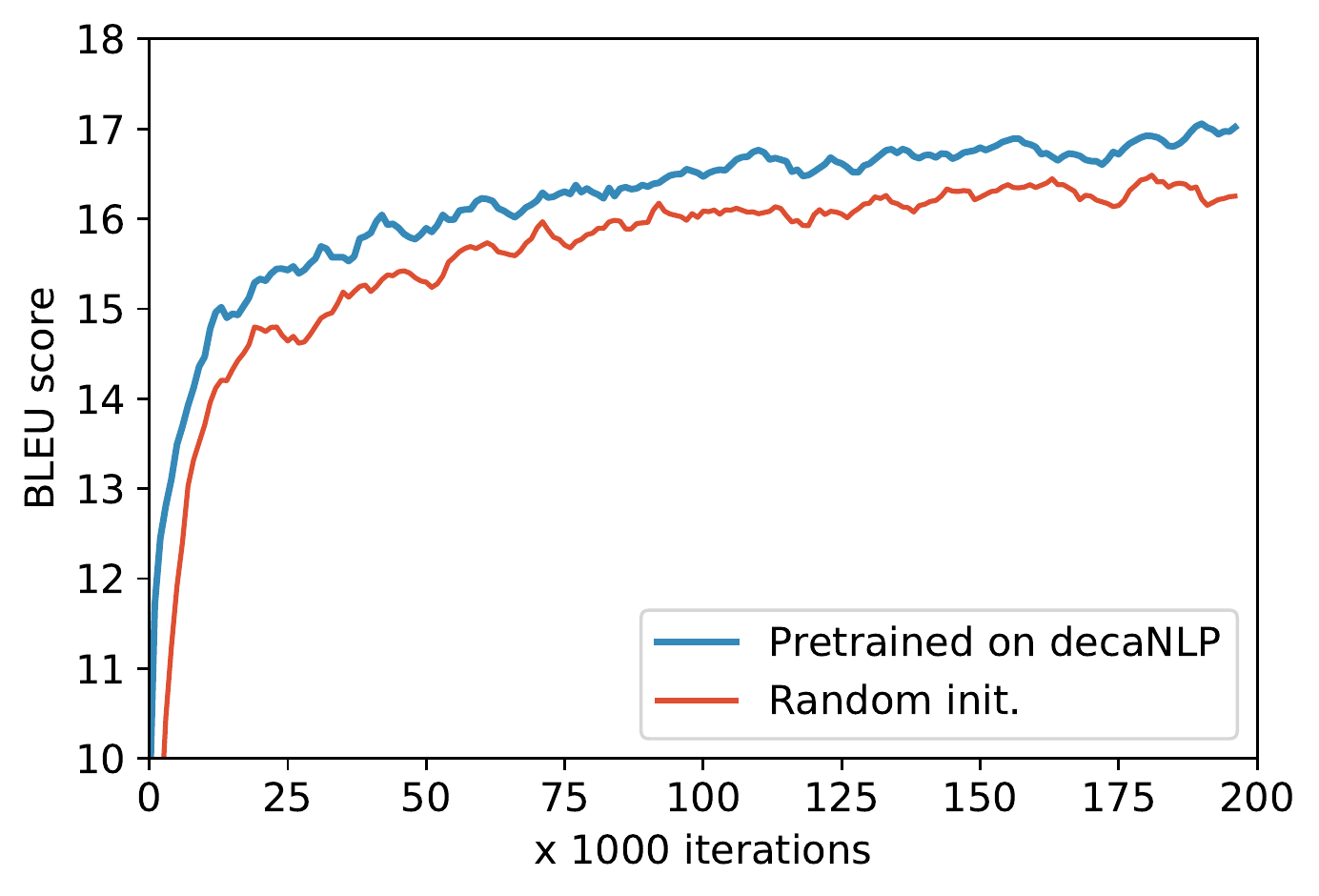}
\end{subfigure}
\qquad
\begin{subfigure}[t]{0.45\textwidth}\label{fig:betterInitializationNER}
\includegraphics[width=\linewidth]{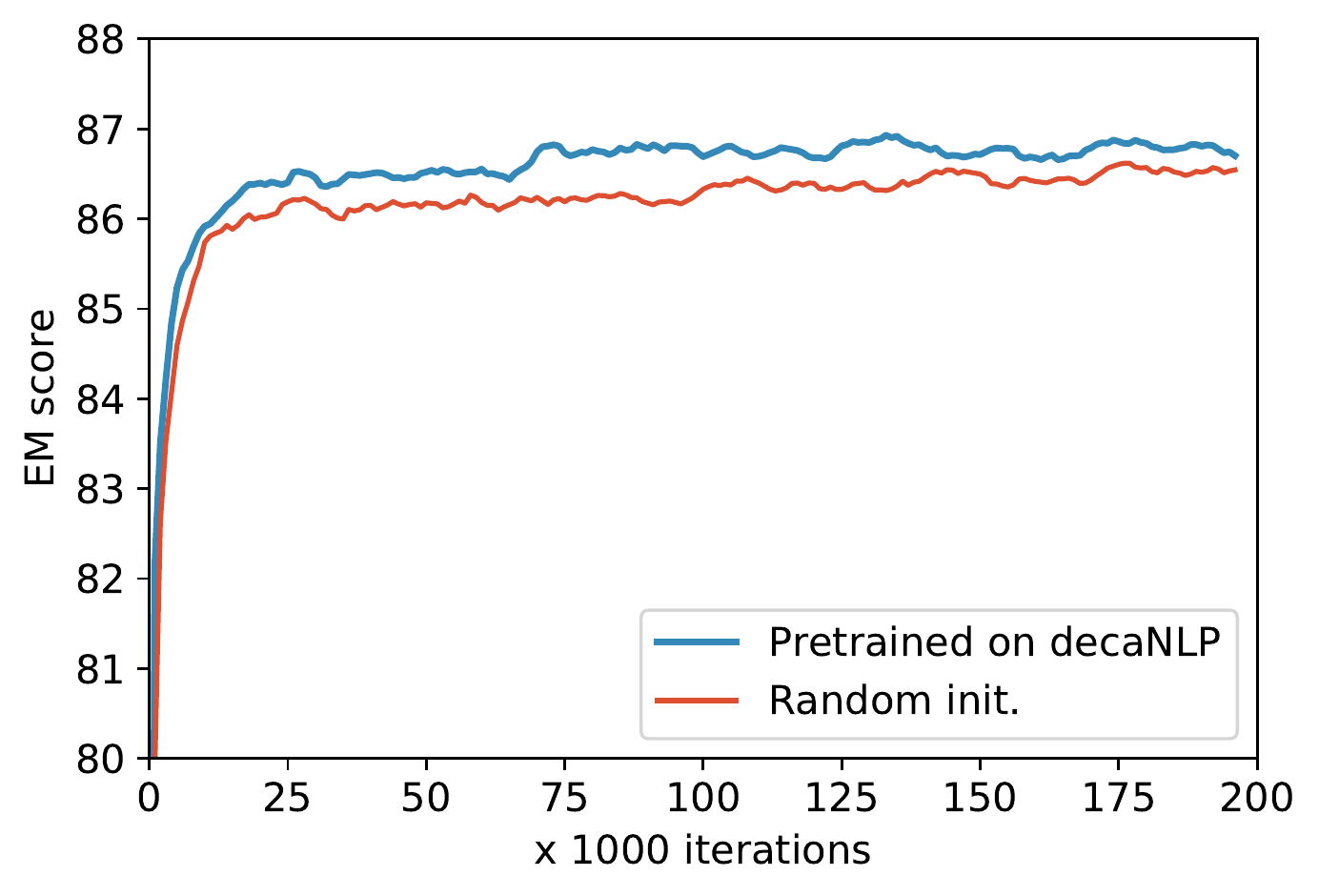}
\end{subfigure}
\caption{\mdl~pretrained on \chal~outperforms random initialization when adapting to new domains and learning new tasks. Left: training on a new language pair -- English to Czech, right: training on a new task -- Named Entity Recognition (NER).} \label{fig:betterInitialization}
\end{figure}

For SST, MNLI, and MWSC, the model prefers the question pointer because the question contains the tokens for acceptable classes. Because the model learns to use the question pointer in this way, it can do zero-shot classification as discussed in~\ref{zeroShot}. For IWSLT and WOZ, the model prefers generating from the vocabulary because German words and dialogue state fields are rarely in the context. The models also avoids copying for QA-ZRE; half of those examples require generating `unanswerable' from the external vocabulary. 

Sampled answers confirm that the model does not confuse tasks. German words are only ever output during translation from English to German. The model never outputs anything but 'positive' and 'negative' for sentiment analysis.

\textbf{Adaptation to new tasks.}
\mdl~trained on\chal~learn to generalize beyond the specific domains for any one task while also learning representations that make learning completely new tasks easier. 
For two new tasks (English-to-Czech translation and named entity recognition - NER), fine-tuning a \mdl~trained on \chal~requires fewer iterations and reaches a better final performance than training from a random initialization (Fig.~\ref{fig:betterInitialization}). For the translation experiment, we use the IWSLT 2016 En$\rightarrow$Cs dataset and for NER, we use OntoNotes 5.0~\citep{Hovy2006OntoNotesT9}.

\textbf{Zero-shot domain adaptation for text classification.}
\label{zeroShot}
Because MNLI is included in \chal, it is possible to adapt to the related Stanford Natural Language Inference Corpus (SNLI)~\citep{Bowman2015ALA}. Fine-tuning a \mdl~pretrained on \chal~achieves an $87\%$ exact match score, which is a $2$ point increase over training from a random initialization and $2$ points from the state of the art~\citep{kim2018semantic}. More remarkably, without any fine-tuning on SNLI, a \mdl~pretrained on \chal~still achieves an exact match score of $62\%$.

Because \chal~contains SST, it can also perform well on other binary sentiment classification tasks. 
On Amazon and Yelp reviews~\citep{amznyelp}, a \mdl~pretrained on \chal~achieves exact match scores of $82.1\%$ and $80.8\%$, respectively, without any fine-tuning.

Additionally, rephrasing questions by replacing the tokens for the training labels \textit{positive/negative} with \textit{happy/angry} or \textit{supportive/unsupportive} at inference time, leads to only small degradation in performance. The model's reliance on the question pointer for SST (see Figure~\ref{pointerUsage}) allows it to copy different, but related class labels with little confusion. This suggests these multitask models are more robust to slight variations in questions and tasks and can generalize to new and unseen classes.

These results demonstrate that models trained on \chal~have potential to simultanesouly generalize to out-of-domain contexts and questions for multiple tasks and even adapt to unseen classes for text classification. This kind of zero-shot domain adaptation in both input and output spaces suggests that the breadth of tasks in \chal~encourages generalization beyond what can be achieved by training for a single task.

\section{Related Work}
This section contains work related to aspects of \chal~and \mdl~that are not task-specific. See Appendix~\ref{taskSpecifcRelatedWork} for work related to each individual task.
\paragraph{Transfer Learning in NLP.}
Most success in making use of the relatedness between natural language tasks stem from transfer learning. 
Word2Vec~\citep{Mikolov2013EfficientEO,Mikolov2013DistributedRO}, skip-thought vectors~\citep{Kiros2015SkipThoughtV} and GloVe~\citep{Pennington2014GloveGV} yield pretrained embeddings that capture useful information about natural language. 
The embeddings~\citep{Collobert2008AUA,Collobert2011NaturalLP}, intermediate representations~\citep{Peters2018DeepCW}, and weights of language models can be transferred to similar architectures~\citep{Ramachandran2017UnsupervisedPF} and classification tasks~\citep{Howard2018FinetunedLM}.
Intermediate representations from supervised machine translation models improve performance on question answering, sentiment analysis, and natural language inference~\citep{McCann2017LearnedIT}. 
Question answering datasets support each other as well as entailment tasks~\citep{Min2017QuestionAT}, and high-resource machine translation can support low-resource machine translation~\citep{Zoph2016TransferLF}. 
This work shows that the combination of \mdl~and \chal~makes it possible to transfer an entire end-to-end model that can be adapted for any NLP task cast as question answering.

\paragraph{Multitask Learning in NLP.}
Unified architectures have arisen for chunking, POS tagging, NER, and SRL~\citep{Collobert2011NaturalLP} as well as dependency parsing, semantic relatedness, and natural language inference~\citep{Hashimoto2016AJM}.
Multitask learning over different machine translation language pairs can enable zero-shot translation~\citep{Johnson2017GooglesMN},
and sequence-to-sequence architectures can be used to multitask across translation, parsing, and image captioning~\citep{Luong2015MultitaskST} using varying numbers of encoders and decoders.
These tasks can also be learned with image classification and speech recognition with careful modularization~\citep{Kaiser2017OneMT}, and the success of this approach extends to visual and textual question answering~\citep{Xiong2016}. 
Learning such modularization can further mitigate interference between tasks~\citep{Ruder2017SluiceNL}.

More generally, multitask learning has been successful when models are able to capitalize on relatedness amongst tasks while mitigating interference from dissimilarities~\citep{Caruana1997MultitaskL}. When tasks are sufficiently related, they can provide an inductive bias~\citep{Mitchell1980TheNF} that forces models to learn more generally useful representations. By unifying tasks under a single perspective, it is possible to explore these relationships~\citep{wang2018glue,poliak2018evaluation,poliak2018towards}.

\mdl~trained on \chal~is the first, single model to achieve reasonable performance on such a wide variety of complex NLP tasks without task-specific modules or parameters, with little evidence of catastrophic interference, and without parse trees, chunks, POS tags, or other intermediate representations. This sets the foundation for general question answering models.

\paragraph{Optimization and Catastrophic Forgetting.}
Multitask learning presents a set of optimization problems that extend beyond the NLP setting. Multi-objective optimization \citep{deb2014multi} naturally connects to multitask learning and typically involves querying a decision-maker who weighs different objectives. 
Much effort has gone into mitigating catastrophic forgetting~\citep{McCloskey1989CatastrophicII,Ratcliff1990ConnectionistMO,Kemker2017MeasuringCF} by penalizing the norm of parameters when training on a new task \citep{Kirkpatrick2017OvercomingCF}, the norm of the difference between parameters for previously learned tasks during parameter updates~\citep{Hashimoto2016AJM}, incrementally matching modes \citep{lee2017overcoming}, rehearsing on old tasks \citep{robins1995catastrophic}, using adaptive memory buffers \citep{gepperth2016bio}, finding task-specific paths through networks \citep{fernando2017pathnet}, and packing new tasks into already trained networks~\citep{mallya2017packnet}.

\mdl~is able to perform nearly as well or better in the multitask setting as in the single-task setting for each task despite being capped at the same number of trainable parameters in both. A collection of {\mdl}s trained for each task individually would use far more trainable parameters than a single \mdl~trained jointly on \chal. This suggests that MQAN successfully uses trainable parameters more efficiently in the multitask setting by learning to pack or share parameters in a way that limits catastrophic forgetting.

\paragraph{Meta-Learning}
Meta-learning attempts to train models on a variety of tasks so that they can easily learn new tasks~\citep{thrun1998learning,thrun1998lifelong,vilalta2002perspective}. Past work has shown how to learn rules for learning~\citep{schmidhuber1987evolutionary,Bengio1992}, train meta-agents that control parameter updates~\citep{hochreiter2001learning,andrychowicz2016learning}, augment models with special memory mechanisms~\citep{santoro2016meta,schmidhuber1992learning}, and maximize the degree to which models can learn new tasks~\citep{Finn2017}.

\section{Conclusion}

We introduced the \challenge~(\chal), a new benchmark for measuring the performance of NLP models across ten tasks that appear disparate until unified as question answering.
We presented~\mdl, a model for general question answering that uses a multi-pointer-generator decoder to capitalize on questions as natural language descriptions of tasks. Despite not having any task-specific modules, we trained \mdl~on all \chal~tasks jointly, and we showed that anti-curriculum learning gave further improvements.
After training on \chal~, \mdl~exhibits transfer learning and zero-shot capabilities. When used as pretrained weights, \mdl~improved performance on new tasks. It also demonstrated zero-shot domain adaptation capabilities on text classification from new domains. We hope the the decaNLP benchmark, experimental results, and publicly available code encourage further research into general models for NLP.

\bibliographystyle{plain}
\bibliography{references}
\appendix
\section{Further Related Work}
\label{taskSpecifcRelatedWork}
\textbf{Question Answering.}
Early success on the SQuAD dataset exploited the fact that all answers can be found verbatim in the context. State-of-the-art models point to start and end tokens in the document~\citep{Seo2017BidirectionalAF, Xiong2016DynamicCN,Yu2016EndtoEndRC,Weissenborn2017MakingNQ}. This allowed deterministic answer extraction to overtake sequential token generation~\citep{wang2017machine}. 
This quirk of the dataset does not hold for question answering in general, so recent models for SQuAD are not necessarily general question answering models~\citep{Yu2018QanetCL,Hu2017ReinforcedMR,Wang2017GatedSN,Liu2017StochasticAN,Huang2017FusionNetFV,Xiong2017DCNMO,Liu2017PhaseCO,Pan2017MEMENME,Salant2017ContextualizedWR}. 
While datasets like TriviaQA~\citep{Joshi2017TriviaQAAL} and NewsQA~\citep{Trischler2017NewsQAAM} could also represent question answering, SQuAD is particularly interesting because the human level performance of SQuAD models in the single-task setting depends on a quirk that does not generalize to all forms of question answering. Including SQuAD in \chal~challenges models to integrate techniques learned from a single-task approach into a more general approach while evaluation remains grounded in the document. Many of the alternatives are larger and can be used as additional training data or incorporated into future iterations of the \chal~once the more well-understood SQuAD dataset has been mastered in the multitask setting.

\textbf{Machine Translation.}
Until recently, the standard approach trained recurrent models with attention~\citep{Luong2015EffectiveAT,Bahdanau2014NeuralMT} on a single source-target language pair~\citep{Wu2016GooglesNM,Sennrich2017TheUO}. 
Models that use only convolution~\citep{Gehring2017ConvolutionalST} or attention~\citep{Vaswani2017AttentionIA} have shown that recurrence is not essential for the task, but recurrence can contribute to the strongest models~\citep{chen2018best}. 
While training these models on many source and target languages at the same time remains difficult, limiting models to one source language and many target languages or vice versa can lead to strong performance when resources are limited or null~\citep{Johnson2017GooglesMN}.

While much larger corpora and many other language pairs exist, the English-German IWSLT dataset provides the same order of magnitude of training data as the other tasks in \chal. We encourage the use of larger corpora or multiple language pairs to improve performance, but we did not want to skew the first iteration of the challenge too far towards machine translation.

\paragraph{Summarization}
Recent approaches combine recurrent neural networks with pointer networks to generate output sequences that contain key words copied from the document~\citep{Nallapati2016AbstractiveTS}.
Coverage mechanisms~\citep{Nallapati2016AbstractiveTS,See2017GetTT,Suzuki2017CuttingoffRR} and temporal attention~\citep{paulus2017deep} improve problems with redundancy in long summaries.
Reinforcement learning has pushed performance using common summarization metrics~\citep{paulus2017deep} as well as alternative metrics that transfer knowledge from another task~\citep{Pasunuru2017TowardsIA,Pasunuru2018MultiRewardRS}.

While new corpora like NEWSROOM~\citep{Grusky2018Newsroom} are even larger, CNN/DM remains the current standard benchmark, so we include it in \chal~and encourage augmentation with datasets like NEWSROOM.

\paragraph{Natural Language Inference} 
NLI has a long history playing roles in tasks like information retrieval and semantic parsing~\citep{Fyodorov2000ANL,Condoravdi2003EntailmentIA,Bos2005RecognisingTE,Dagan2005ThePR,MacCartney2009AnEM}.
The introduction of the Stanford Natural Language Inference Corpus (SNLI) by~\citep{Bowman2015ALA} spurred a new wave of interest in NLI, its connections to other tasks, and general sentence representations.
The most successful approaches make use of attentional models that match and align words in the premise to those in the hypothesis~\citep{Tay2017ACA,Peters2018DeepCW,Ghaeini2018DRBiLSTMDR,Chen2017NaturalLI,Wang2017BilateralMM,McCann2017LearnedIT},
but recent non-attentional models designed to extract useful sentence representations have nearly closed the gap~\citep{Liu2017StochasticAN,Im2017DistancebasedSN,Shen2018ReinforcedSN,Choi2017LearningTC}.

The dataset we use, the Multi-Genre Natural Language Inference Corpus (MNLI) introduced by~\citep{Williams2017ABC}, is the successor to SNLI.
Recent approaches to MNLI use methods developed on SNLI and have even pointed out the similarities between models for question answering and NLI~\citep{Huang2017FusionNetFV}.

\paragraph{Sentiment Analysis}
Because SST came with parse trees for every example,
some approaches use all of the sub-tree labels by modeling trees explicitly~\citep{Yu2017NeuralTI,Tai2015ImprovedSR} as in the original paper.
Others use sub-tree labels implicitly~\citep{Yu2017NeuralSE,McCann2017LearnedIT,Peters2018DeepCW},
and still others do not use the sub-trees at all~\citep{Radford2017LearningTG}.
This suggests that while the many sub-tree labels might facilitate learning, they are not necessary to train state-of-the-art models.

\paragraph{Semantic Role Labeling}
Traditionally, models have made use of syntactic parsing information~\cite{Punyakanok2008TheIO}, but recent methods have demonstrated that it is not necessary to use syntactic information as additional input~\citep{Zhou2015EndtoendLO,Marcheggiani2017ASA}. 
State-of-the-art approaches treat SRL as a tagging problem~\citep{He2017DeepSR}, make use of that specific structure to constrain decoding, and mix recurrent and self-attentive layers~\citep{Tan2017DeepSR}.

Because QA-SRL treats SRL as question answering~\citep{He2015QuestionAnswerDS}, it abstracts away the many task-specific constraints of treating SRL as a tagging problem with hand-designed verb-specific roles or grammars. This preserves much of the structure extracted by prior formulations while also allowing models to extract structure that is not syntax-based.

\paragraph{Relation Extraction}
QA-ZRE introduced a similar idea for relation extraction~\citep{Levy2017ZeroShotRE}. By associating natural language questions with relations, this dataset reduces relation extraction to question answering. This makes it possible to use question answering models in place of more traditional relation extraction models that often do not make use of the linguistic similarities amongst relations. This in turn makes it possible to do zero-shot relation extraction.

\paragraph{Goal-Oriented Dialogue}

Dialogue state tracking requires a system to estimate a users goals and and requests given the dialogue context, and it plays a crucial role in goal-oriented dialogue systems. Most models use a structured approach~\citep{mrkvsic2016neural}, with the most recent work making use of both global and local modules to learns representations of the user utterance and previous system actions~\citep{zhong2018global}.

\paragraph{Semantic Parsing}
Similarly, recent approaches to the semantic parsing WikiSQL dataset have made use of structured approaches that move from coarse sketches of the input to fine-grained structured outputs~\citep{dong2018coarse}, direclty employing a type system~\citep{yu2018typesql}, or making use of dependency graphs~\citep{huang2018natural}. 
\section{Preprocessing and Training Details}
\label{preprocessingTrainingDetails}
All data is lowercased as is common for SQuAD, IWSLT, CNN/DM, and WikiSQL; casing is irrelevant for the evaluation of the other tasks.
We use the RevTok tokenizer\footnote{\url{https://github.com/jekbradbury/revtok}} to provide simple, yet completely reversible tokenization,
which is crucial for detokenizing generated sequences for evaluation.
The generative vocabulary in Eq.~\ref{generativeVocabulary} contains the most frequent $50000$ words in the combined training sets for all tasks in \chal.
SQuAD examples with context longer than $400$ tokens were excluded during training and CNN/DM examples had contexts truncated to $400$ tokens during training and evaluation.
Only MNLI examples with a label other than `-' were included during training and evaluation as is standard.
For WOZ, we train turn-by-turn to predict the change in belief state including user requests as an additional slot, but during evaluation we only consider the cumulative belief state as is standard.
We do not perform any form of beam search or otherwise refine greedily sampled outputs for any tasks to avoid task-specific post-processing where possible.

The \mdl~defined in Section~\ref{model} takes $300$-dimensional GloVe embeddings trained on CommonCrawl~\citep{Pennington2014GloveGV} as input.
Words that do not have corresponding GloVe embeddings are assigned zero vectors instead. 
We concatenate $100$-dimensional character n-gram embeddings~\citep{Hashimoto2016AJM} to the GloVe embeddings. 
This corresponds to setting $d_{emb}=400$ in Section~\ref{model}.
Internal model dimension $d=200$, hidden dimension $f=150$, and the number of heads in multi-head attention $p=3$.
\mdl~uses $2$ self-attention and multi-head decoder attention layers. 
We use a dropout of $0.2$ on inputs to LSTMs, layers following coattention, and decoder layers, before multiplying by $\tilde Z$ in Eq.~\ref{attention}, before adding $X$ in Eq.~\ref{ffn}, and generally after any linear transformation. 
The models are trained using Adam with $(\beta_1, \beta_2, \epsilon) = (0.9, 0.98, 10^{-9})$ and a warmup schedule~\citep{Vaswani2017AttentionIA}, which increases the learning rate linearly from $0$ to $2.5 \times 10^{-3}$ over $800$ iterations before decaying it as $\frac{1}{\sqrt{k}}$, where $k$ is the iteration count. Batches consist entirely of examples from one task and are dynamically constructed to fit as many examples as possible so that the sum of the number of tokens in the context and question and five times the number of tokens in the asnwer does not exceed $10000$.
\section{\Model~(\mdl) Encoder}\label{sec:encoderDetails}
Recall from Section~\ref{model} that the encoder has three input sequences during training: a context $c$ with $l$ tokens, a question $q$ with $m$ tokens, and an answer $a$ with $n$ tokens. Each of these is represented by a matrix where the $i$th row of the matrix corresponds to a $d_{emb}$-dimensional embedding (such as word or character vectors) for the $i$th token in the sequence: \begin{equation}
C \in \mathbb{R}^{l \times d_{emb}}
\qquad 
Q \in \mathbb{R}^{m \times d_{emb}}
\qquad A \in \mathbb{R}^{n \times d_{emb}}
\end{equation}

\textbf{Independent Encoding.} A linear layer projects input matrices onto a common $d$-dimensional space. 
\begin{equation}
CW_1 = C_{proj} \in \mathbb{R}^{l \times d} 
\qquad 
QW_1 = Q_{proj} \in \mathbb{R}^{m \times d} 
\end{equation}
These projected representations are fed into a shared, bidirectional Long Short-Term Memory Network (BiLSTM)
\citep{Hochreiter1997LongSM,Graves2005FramewisePC} \footnote{
For input $X\in \mathbb{R}^{T\times d_{in}}$, let $h^{\forward}_t=\lstm{x^T_t}{h^{\forward}_{t-1}}$ 
and 
$h^{\backward}_t = \lstm{x^T_t}{h^{\backward}_{t+1}}$.
Representations are concatenated along the last dimension $h_t=\left[ h^{\forward}_t; h^{\backward}_t \right]$ for each $t$ and stacked as rows of output $H\in \mathbb{R}^{T\times d_{out}}$.}
\begin{equation}
\text{BiLSTM}_{ind}(C_{proj}) = C_{ind}  \in \mathbb{R}^{l \times d}  
\qquad 
\text{BiLSTM}_{ind}(Q_{proj}) = Q_{ind}  \in \mathbb{R}^{m \times d} 
\end{equation}

\textbf{Alignment.} We obtain coattended representations by first aligning encoded representations of each sequence.
We add separate trained, dummy embeddings to $C_{ind}$ and $Q_{ind}$ (now  $\in \mathbb{R}^{(l+1) \times d}$ and $\mathbb{R}^{(m+1) \times d}$) so that tokens are not forced to align with any token in the other sequence.

Let $\softmax{X}$ denote a column-wise softmax that normalizes each column of the matrix $X$ to have entries that sum to $1$. 
We obtain alignments by normalizing dot-product similarity scores between representations of one sequence with those of the other:
\begin{equation}
\softmax{C_{ind}Q_{ind}^\top} = S_{cq} \in \mathbb{R}^{(l+1) \times (m+1)}
\qquad
\softmax{Q_{ind}C_{ind}^\top} = S_{qc} \in \mathbb{R}^{(m+1) \times (l+1)}
\end{equation}

\textbf{Dual Coattention.} These alignments are used to compute weighted summations of the information from one sequence that is relevant to a single token in the other. 
\begin{equation}
S_{cq}^\top C_{ind} = C_{sum} \in \mathbb{R}^{(m+1) \times d}
\qquad
S_{qc}^\top Q_{ind} = Q_{sum} \in \mathbb{R}^{(l+1) \times d}
\end{equation}

The coattended representations use the same weights to transfer information gained from alignments back to the original sequences:
\begin{equation}
S_{qc}^\top C_{sum} = C_{coa} \in \mathbb{R}^{(l+1) \times d}
\qquad
S_{cq}^\top Q_{sum} = Q_{coa} \in \mathbb{R}^{(m+1) \times d}
\end{equation}
The first column of the summation and coattentive representations correspond to the dummy embeddings. This information is not needed, so we drop that column of the matrices to get $C_{coa} \in \mathbb{R}^{l \times d}$ and $Q_{coa} \in \mathbb{R}^{m \times d}$.

\textbf{Compression.} In order to compress information from dual coattention back to the more manageable dimension $d$, we concatenate all four prior representations for each sequence along the last dimension and feed into separate BiLSTMs: \begin{equation}
\text{BiLSTM}_{comC}([C_{proj}; C_{ind}; Q_{sum}; C_{coa}]) = C_{com}  \in \mathbb{R}^{l \times d}
\end{equation}
\begin{equation}
\text{BiLSTM}_{comQ}([Q_{proj}; Q_{ind}; C_{sum}; Q_{coa}]) = Q_{com}  \in \mathbb{R}^{m \times d}
\end{equation}

\textbf{Self-Attention.} Next, we use multi-head, scaled dot-product attention~\citep{Vaswani2017AttentionIA} to capture long distance dependencies within each sequence. Let
\begin{equation}
\text{Attention}(\tilde X,\tilde Y,\tilde Z) = \text{softmax}\left(\frac{\tilde X \tilde Y^\top}{\sqrt{d}}\right) \tilde Z
\label{attention}
\end{equation}
\begin{equation}
\label{multihead}
\text{MultiHead}(X, Y, Z) = [h_1;\cdots;h_p]W_o \qquad
\text{where } h_j = \text{Attention}(XW_j^X, YW_j^Y, ZW_j^Z)
\end{equation}
All linear transformations in Eq.~\eqref{multihead} project to $d$ so that multi-head attention representations maintain dimensionality:
\begin{equation}
\text{MultiHead}_C(C_{com}, C_{com}, C_{com}) = C_{mha}
\qquad
\text{MultiHead}_Q(Q_{com}, Q_{com}, Q_{com}) = Q_{mha}
\end{equation}

We then use projected, residual feedforward networks (FFN) with ReLU activations~\citep{Nair2010RectifiedLU,Vaswani2017AttentionIA} and layer normalization~\citep{Ba2016LayerN} on the inputs and outputs. With parameters $U\in \mathbb{R}^{d \times f}$ and $V\in \mathbb{R}^{f \times d}$:
\begin{equation}
\label{ffn}
FFN(X) = \text{max}(0, XU)V + X
\end{equation}
\begin{equation}
FFN_C(C_{com} + C_{mha}) = C_{self}  \in \mathbb{R}^{l \times d}
\qquad
FFN_Q(Q_{com} + Q_{mha}) = Q_{self}  \in \mathbb{R}^{m \times d}
\end{equation}

\textbf{Final Encoding.} Finally, we aggregate all of this information across time with two BiLSTMs:
\begin{equation}
\text{BiLSTM}_{finC}(C_{self}) = C_{fin}  \in \mathbb{R}^{l \times d}
\qquad
\text{BiLSTM}_{finQ}(Q_{self}) = Q_{fin}  \in \mathbb{R}^{m \times d}
\end{equation}

These matrices are given to the decoder to generate the answer.

\section{Curriculum Learning}
\label{appendix:curriculum}
\begin{table}[b]
\caption{Validation metrics for \mdl~using various training strategies. The first is fully joint, which samples batches round-robin from all tasks. Others first use a curriculum or anti-curriculum schedule over a subset of tasks before switching to fully joint over all tasks. Curriculum first trains tasks that take relatively few iterations to converge when trained alone. This omits SQuAD, IWSLT, CNN/DM, and MNLI. The remaining strategies are anti-curriculum. They include in the first phase either SQuAD alone, SQuAD, IWSLT, and CNN/DM, or SQuAD, IWSLT, CNN/DM, and MNLI. \label{curriculumAblations}}
\centering
\begin{tabular}{lrrrrr}\toprule
&&&\multicolumn{3}{c}{Anti-Curriculum}\\
\cmidrule[.5pt](lr){4-6}
Dataset  & Fully Joint
		 & Curriculum
         & SQuAD
         & +IWSLT+CNN/DM 
         & +MNLI\\
         \midrule
SQuAD  & 70.8 & 43.4 & 74.3 &  74.5 & 74.6 \\    
IWSLT  & 16.1 & 4.3 & 13.7 &  18.7 &   19.0 \\ 
CNN/DM & 23.9 & 21.3 & 24.6 & 20.8  &    21.6 \\ 
MNLI   & 70.5 & 58.9 & 69.2 & 69.6 &  72.7 \\ 
SST    & 86.2 & 84.5 & 86.4 & 83.6 & 86.8 \\ 
QA-SRL & 75.8 & 70.6 & 77.6 & 77.5 &   75.1 \\ 
QA-ZRE & 28.0 & 24.6 & 34.7 & 30.1 &   37.7 \\ 
WOZ    & 80.6 & 81.9 & 84.1 & 81.7 &   85.6 \\ 
WikiSQL& 62.0 & 68.6 & 58.7 & 54.8 &  42.6 \\ 
MWSC   & 48.8 & 41.5 & 48.4 & 34.9 & 41.5 \\ 
\midrule
decaScore &562.7&	499.6&	571.7	&546.2&	557.2\\
\bottomrule\end{tabular}\end{table}

For multitask training, we experiment with various round-robin batch-level sampling strategies.

The first strategy we consider is fully joint. In this strategy, batches are sampled round-robin from all tasks in a fixed order from the start of training to the end. This strategy performed well on tasks that required fewer iterations to converge during single-task training (see Table~\ref{curriculumAblations}), but the model struggles to reach single-task performance for several other tasks. In fact, we found a correlation between the performance gap between single and multitasking settings of any given task and number of iterations required for convergence for that task in the single-task setting. 

With this in mind, we experimented with several anti-curriculum schedules \cite{Bengio2009CurriculumL}. These training strategies all consist of two phases. In the first phase, only a subset of the tasks are trained jointly, and these are typically the ones that are more difficult. In the second phase, all tasks are trained according to the fully joint strategy.

We first experimented with isolating SQuAD in the first phase, and the switching to fully joint training over all tasks. Since we take a question answering approach to all tasks, we were motivated by the idea of pretraining on SQuAD before being exposed to other kinds of question answering. This would teach the model how to use the multi-context decoder to properly retrieve information from the context before needing to learn how to switch between tasks or generate words on its own. Additionally, pretraining on SQuAD had already been shown to improve performance for NLI~\citep{Min2017QuestionAT}. Empirically, we found that this motivation is well-placed and that this strategy outperforms all others that we considered in terms of the decaScore. This strategy sacrificed performance on IWSLT but recovered the lost decaScore on other tasks, especially those which use pointers. 

To explore if adding additional tasks to the initial curriculum would improve performance further, we experimented with adding IWSLT and CNN/DM to the first phase and in another experiment, adding IWSLT, CNN/DM and MNLI. These are tasks with a large number of training examples relative to the other tasks, and they contain the longest answer sequences. Further, they form a diverse set since they encourage the model to decode in different ways such as the vocabulary for IWSLT, context-pointer for SQuAD and CNN/DM, and question-pointer for MNLI. In our results, we however found no improvement by adding these tasks. In fact, in the case when we added SQuAD, IWSLT, CNN/DM and MNLI to the initial curriculum, we observed a marked degradation in performance of some other tasks including QA-SRL, WikiSQL and MWSC. This suggests that it is concordance between the question answering nature of the task and SQuAD that enabled improved outcomes and not necessarily the richness of the task.

Finally, as a check to our hypothesis, we also tried a curriculum schedule that used SST, QA-SRL, QA-ZRE, WOZ, WikiSQL and MWSC in the initial curriculum. This effectively takes the easiest tasks  and trains on those first. This was indubitably an inferior strategy; not only does the model perform worse on tasks that were not in the initial curriculum, especially SQuAD and IWSLT, it also performs worse on the tasks that were. Finding that anti-curriculum learning benefited models in the \chal~also validated intuitions outlined in~\citep{Caruana1997MultitaskL}: tasks that are easily learned may not lead to development of internal representations that are useful to other tasks. Our results actually suggest a stronger claim: including easy tasks early on in training makes it more difficult to learn internal representations that are useful to other tasks.

We note in passing that the results above underscores the challenges and trade-offs in the multitasking setting. By ordering the tasks differently, it is possible to improve performance on some of the tasks but that improvement is not without a concomitant drop in performance for others. Indeed, a gap still exists between single-task performance and the results above. The question of how this gap can be bridged is a topic of continued research. 
\end{document}